\PassOptionsToPackage{unicode}{hyperref}
\PassOptionsToPackage{hyphens}{url}
\PassOptionsToPackage{dvipsnames,svgnames,x11names}{xcolor}

\documentclass[12pt]{article}
\usepackage[utf8]{inputenc}
\usepackage{amsmath,amsthm,amssymb,amsfonts}

\usepackage{algorithm}
\usepackage{algpseudocode}
\usepackage{enumitem}
\usepackage{multirow}

\usepackage{indentfirst}
\usepackage{graphicx}
\graphicspath{{Figures/}}
\usepackage{subcaption}
\usepackage{xr-hyper}

\newtheorem{theorem}{Theorem}

\newtheorem{corollary}{Corollary}
\newtheorem{proposition}{Proposition}
\newtheorem{assumption}{Assumption}

\usepackage[section]{placeins}  
\DeclareFontFamily{U}{mathx}{}
\DeclareFontShape{U}{mathx}{m}{n}{<-> mathx10}{}
\DeclareSymbolFont{mathx}{U}{mathx}{m}{n}
\DeclareMathAccent{\widehat}{0}{mathx}{"70}
\DeclareMathAccent{\check}{0}{mathx}{"71}

\usepackage{iftex}
\ifPDFTeX
  \usepackage[T1]{fontenc}
  \usepackage[utf8]{inputenc}
  \usepackage{textcomp} 
\else 
  \usepackage{unicode-math}
  \defaultfontfeatures{Scale=MatchLowercase}
  \defaultfontfeatures[\rmfamily]{Ligatures=TeX,Scale=1}
\fi
\usepackage{lmodern}
\ifPDFTeX\else  
\fi
\IfFileExists{upquote.sty}{\usepackage{upquote}}{}
\IfFileExists{microtype.sty}{
  \usepackage[]{microtype}
  \UseMicrotypeSet[protrusion]{basicmath} 
}{}
\makeatletter
\@ifundefined{KOMAClassName}{
  \IfFileExists{parskip.sty}{%
    \usepackage{parskip}
  }{
    \setlength{\parindent}{0pt}
    \setlength{\parskip}{6pt plus 2pt minus 1pt}}
}{
  \KOMAoptions{parskip=half}}
\makeatother
\usepackage{xcolor}
\makeatletter
\ifx\paragraph\undefined\else
  \let\oldparagraph\paragraph
  \renewcommand{\paragraph}{
    \@ifstar
      \xxxParagraphStar
      \xxxParagraphNoStar
  }
  \newcommand{\xxxParagraphStar}[1]{\oldparagraph*{#1}\mbox{}}
  \newcommand{\xxxParagraphNoStar}[1]{\oldparagraph{#1}\mbox{}}
\fi
\ifx\subparagraph\undefined\else
  \let\oldsubparagraph\subparagraph
  \renewcommand{\subparagraph}{
    \@ifstar
      \xxxSubParagraphStar
      \xxxSubParagraphNoStar
  }
  \newcommand{\xxxSubParagraphStar}[1]{\oldsubparagraph*{#1}\mbox{}}
  \newcommand{\xxxSubParagraphNoStar}[1]{\oldsubparagraph{#1}\mbox{}}
\fi
\makeatother

\usepackage{longtable,booktabs,array}
\usepackage{calc} 
\usepackage{etoolbox}
\makeatletter
\patchcmd\longtable{\par}{\if@noskipsec\mbox{}\fi\par}{}{}
\makeatother
\IfFileExists{footnotehyper.sty}{\usepackage{footnotehyper}}{\usepackage{footnote}}
\makesavenoteenv{longtable}
\usepackage{graphicx}
\makeatletter
\def\maxwidth{\ifdim\Gin@nat@width>\linewidth\linewidth\else\Gin@nat@width\fi}
\def\maxheight{\ifdim\Gin@nat@height>\textheight\textheight\else\Gin@nat@height\fi}
\makeatother
\setkeys{Gin}{width=\maxwidth,height=\maxheight,keepaspectratio}
\makeatletter
\def\fps@figure{htbp}
\makeatother

\makeatletter
\@ifpackageloaded{caption}{}{\usepackage{caption}}
\AtBeginDocument{%
\ifdefined\contentsname
  \renewcommand*\contentsname{Table of contents}
\else
  \newcommand\contentsname{Table of contents}
\fi
\ifdefined\listfigurename
  \renewcommand*\listfigurename{List of Figures}
\else
  \newcommand\listfigurename{List of Figures}
\fi
\ifdefined\listtablename
  \renewcommand*\listtablename{List of Tables}
\else
  \newcommand\listtablename{List of Tables}
\fi
\ifdefined\figurename
  \renewcommand*\figurename{Figure}
\else
  \newcommand\figurename{Figure}
\fi
\ifdefined\tablename
  \renewcommand*\tablename{Table}
\else
  \newcommand\tablename{Table}
\fi
}
\@ifpackageloaded{float}{}{\usepackage{float}}
\floatstyle{ruled}
\@ifundefined{c@chapter}{\newfloat{codelisting}{h}{lop}}{\newfloat{codelisting}{h}{lop}[chapter]}
\floatname{codelisting}{Listing}

\makeatother
\makeatletter
\@ifpackageloaded{caption}{}{\usepackage{caption}}
\@ifpackageloaded{subcaption}{}{\usepackage{subcaption}}
\makeatother

\ifLuaTeX
  \usepackage{selnolig}  
\fi
\usepackage[]{natbib}
\usepackage{bookmark}

\IfFileExists{xurl.sty}{\usepackage{xurl}}{} 
\hypersetup{
  pdftitle={Title},
  pdfauthor={Author 1; Author 2},
  pdfkeywords={3 to 6 keywords, that do not appear in the title},
  colorlinks=true,
  linkcolor={blue},
  filecolor={Maroon},
  citecolor={Blue},
  urlcolor={Blue},
  pdfcreator={LaTeX via pandoc}}

\newcommand{\anon}{1}

\begin{document}
\def\spacingset#1{\renewcommand{\baselinestretch}%
{#1}\small\normalsize} \spacingset{1}

\if1\anon
{ 
  \title{\bf 
  Deep Multitask Learning for Mixed-Type Outcomes with Shared Sparsity}
  \author{Huichao Li \\
    {\small School of Mathematical Sciences, University of Chinese Academy of Sciences}\\
    Tong Wang \\
    {\small School of Statistics and Data Science, Southeast University, Nanjing, Jiangsu, China} \\
    Sanguo Zhang \\
    {\small School of Mathematical Sciences, University of Chinese Academy of Sciences, Beijing, China}\\
    {\footnotesize Key Laboratory of Big Data Mining and Knowledge Management,
Chinese Academy of Sciences, Beijing, China}\\
Shuangge Ma \footnote{Corresponding author: shuangge.ma@yale.edu} \\
{\small Department of Biostatistics, Yale School of Public Health, New Haven, Connecticut, USA}}
    \date{}
  \maketitle
} \fi

\if0\anon
{
  \bigskip
  \bigskip
  \bigskip
  \begin{center}
    {\LARGE\bf Hierarchical Contrastive Learning for Multimodal Data 
    }
\end{center}
  \medskipglo
} \fi

\bigskip
\begin{abstract}
Most existing multitask learning approaches are limited by their reliance on task-specific loss functions tailored to the scale and type of each outcome. When outcomes differ across tasks, these losses are generally not directly comparable, which makes it difficult to formulate a unified objective and may limit information sharing across tasks. We propose a multitask transformation framework in which task-specific responses may differ through unknown monotone transformations. Motivated by high-dimensional biological applications in which the predictor dimension may diverge with the sample size while only a common subset of predictors is informative, we consider shared sparsity across tasks. Under this framework, we estimate the target functions and identify important predictors by optimizing a smoothed rank-based criterion with a group-Lasso penalty, implemented through a multitask deep neural network with a shared first layer. We establish the nonasymptotic excess-risk bounds, and variable-selection consistency for the proposed estimator. Simulation studies show that the proposed method achieves competitive prediction and variable-selection performance compared with competing approaches. Analyses of gene-expression studies with continuous, binary, and mixed outcomes further illustrate that the proposed method improves prediction and identifies biologically meaningful shared predictors.

\end{abstract}

\noindent%

\vfill

\newpage
\spacingset{1.8}

\section{Introduction}
\label{sec:intro}
Multitask learning has become an important area in modern statistical machine learning, as multiple related outcomes are often observed with a common set of predictors. In many applications, these outcomes may differ in scale, type, or observation mechanism, while still depending on a common set of important predictors. For example, genomic and clinical studies often measure continuous biomarkers, binary disease status, and molecular subtypes on the same subjects, leading to related outcomes with different scales but potentially shared molecular predictors. A key objective in such settings is to improve estimation and prediction by leveraging shared information among tasks \citep{zhang2021survey}. Achieving this objective requires not only exploiting task relatedness but also identifying important predictors shared across tasks, since such identification is essential for both accurate estimation and interpretability when outcome types differ across tasks.

Existing statistical approaches to multitask learning have focused primarily on linear frameworks. For example, \citet{huang2025optimal} studied sparse heterogeneity through a global parameter with sparse task-specific deviations and established minimax-optimal procedures for multitask linear regression. \citet{tian2025learning} considered multitask linear regression in which task-specific low-dimensional representations remain close to a central representation. Although these studies provide important statistical insights, they remain largely restricted to linear structure in either the response model or the representation space. Consequently, they do not directly address settings in which task effects are nonlinear and comparability across tasks is affected by differences in scale.

Another line of research uses deep neural networks to enhance the flexibility of multitask modeling, but existing deep multitask methods still rely largely on weighted combinations of task-specific losses. Representative approaches include uncertainty-based weighting \citep{kendall2018multi} and gradient balancing \citep{chen2018gradnorm} for tasks with different scales, as well as architectural approaches that separate shared and task-specific components, such as multi-gate mixture-of-experts \citep{ma2018modeling}. Recent studies have shown that differences in loss scale remain a major obstacle to stable task balancing \citep{qin2025towards}. Although these methods improve flexibility and optimization stability, they still formulate multitask learning through weighted task-specific objectives and therefore do not yield a unified framework for outcomes that differ in scale or type. The central difficulty is therefore not only how to share information across tasks, but also how to define a common estimation target when task-specific losses are intrinsically incomparable.

Theoretical studies of multitask learning have focused largely on linear shared representations or closely related linear structures, including in deep learning settings. For example, \citet{tripuraneni2021provable} studied multitask linear regression with a common low-dimensional representation and derived sample-complexity guarantees for learning and transfer. \citet{tian2025learning} extended this framework by allowing each task to have its own linear representation that deviates from a central representation by a bounded amount. In deep multitask learning, \citet{zakerinia2025low} derived generalization bounds through amortized intrinsic dimensionality under a specific low-dimensional parametrization in which task models are expressed as linear combinations of shared basis elements with task-specific coefficients. The literature on variable selection for multitask learning has also focused primarily on linear structures. For instance, \citet{Lounici2009} studied group-Lasso estimation under shared sparsity and established oracle inequalities and variable-selection properties. \citet{behdin2025multi} relaxed exact sparsity sharing by allowing related tasks to have partially shared sparsity patterns. These approaches do not address the identification of predictors that are jointly informative across tasks with different outcome types under nonlinear models. Thus, a unified statistical framework is still needed to integrate nonlinear estimation, outcomes with different scales or types, and shared variable selection within a single formulation.

In this work, we propose a multitask transformation framework that treats the task-specific target functions as the common objects of estimation while allowing the observed outcomes to differ through unknown monotone transformations. This formulation places different tasks into a unified framework and addresses the incomparability induced by differences in outcome scale or type. Motivated by biological applications with high-dimensional predictors, we consider shared sparsity across tasks, allowing the number of predictors to grow with the sample size while assuming that only a common subset is informative. For feasibility of estimation, we develop a smoothed rank-based objective with a group-Lasso penalty and implement it through a multitask neural network with a shared first layer. We establish the identifiability of the proposed rank-based criterion, nonasymptotic excess-risk bounds, and variable-selection consistency for the proposed estimator. These results provide a unified statistical framework for nonlinear multitask learning with responses that differ in scale or type and with shared sparsity.


\section{Method}\label{sec:method}
\subsection{Multitask Transformation Model}
Suppose that we observe $n$ i.i.d. samples, each consisting of a common predictor vector $\boldsymbol{X}$ and $T$ task-specific responses $(Y_1,\ldots,Y_T)$. Let 
$\{(\boldsymbol{x}_i,y_{1i},\ldots,y_{Ti})\}_{i=1}^n$ denote i.i.d. copies of 
$(\boldsymbol{X},Y_1,\ldots,Y_T)$, where $\boldsymbol{x}_i\in\mathcal{X}\subseteq\mathbb{R}^p$ and $y_{ti}\in\mathbb{R}$ for $t=1,\ldots,T$. We consider the following multitask transformation model
\begin{equation}
Y_t = \mathcal{D}_t\big(f_t^*(\boldsymbol{X})+\epsilon_t\big), \qquad t=1,\ldots,T,
\label{eq:MTL_model}
\end{equation}
where $\mathcal{D}_t:\mathbb{R}\to\mathbb{R}$ is a nondegenerate monotone increasing transformation function, $\boldsymbol{f}^*=(f_1^*,\ldots,f_T^*)$ is the collection of task-specific target functions with $f_t^*:\mathcal{X}\to\mathbb{R}$, and $\epsilon_t$ is a task-specific error independent of $\boldsymbol{X}$ and mutually independent across tasks. The multitask transformation model~\eqref{eq:MTL_model} accommodates response types with different scales or observation mechanisms, including continuous and binary outcomes, among others. This formulation is particularly useful in multitask settings because it treats the target functions $\{f_t^*\}_{t=1}^T$ as the primary targets of estimation, and differences in outcome scale can be absorbed by the monotone transformations $\mathcal{D}_t$.

Motivated by applications where multiple outcomes depend on a common subset of predictors, we consider shared sparsity across tasks. For notational convenience, let the first $p_s$ predictors, $\boldsymbol{X}^s=(X_1,\ldots,X_{p_s})$, be important and the remaining $p_c=p-p_s$ predictors, $\boldsymbol{X}^c=(X_{p_s+1},\ldots,X_p)$, be unimportant. Following \cite{Jean2017} and \citet{FAN2023}, we consider the setting in which $p_s$ is fixed. This assumption induces task relatedness through a common set of important predictors and is standard in linear multitask frameworks \citep{Lounici2009}. It is also appropriate in applications where related outcomes are driven by a common set of predictive features, such as biomarkers or clinical covariates that affect multiple endpoints \citep{Kundu2021}. 

\subsection{Deep Smoothed Rank-based Estimation}
Under model~\eqref{eq:MTL_model}, the primary targets of estimation are the task-specific target functions $f_t^*$. Because the observed responses are linked to the target functions through the task-specific monotone transformations, directly combining conventional task-specific losses can be sensitive to outcome scale and model specification. We therefore consider a rank-based criterion based on within-task pairwise orderings, which is invariant to monotone transformations of the responses. Specifically, the unsmoothed rank-based criterion is
$$ H_{n}(f_t) = \frac{1}{n(n-1)}\sum_{i\neq j} I(y_{ti}>y_{tj}) I\big(f_t(\boldsymbol{x}_i)>f_t(\boldsymbol{x}_j)\big),\qquad t=1,\ldots,T,$$
where $I(\cdot)$ is the indicator function. This criterion measures the agreement between the ordering induced by $f_t$ and the observed ordering of the response, in the spirit of linear maximum rank correlation estimation \citep{Han1987}. To facilitate optimization, we replace the discontinuous indicator function with the smooth approximation $s(\zeta;a_n)=\frac{1}{1+\exp(-\zeta/a_n)}$, where $a_n>0$ is a decreasing sequence satisfying $a_n\to 0$ \citep{Song2007}.

We estimate the target functions using a multitask feedforward neural network $\boldsymbol{f}_{\boldsymbol{\theta}}=(f_{\theta_1},\ldots,f_{\theta_T})^\top$ with a shared first layer, where $\theta_t$ contains the task-specific network parameters for task $t$. To identify important predictors, we consider the penalized smoothed rank-based objective defined as
\begin{equation}\label{eq:penalized_loss} 
\begin{split}
L_n(\boldsymbol{f}_{\boldsymbol{\theta}})
=&-\sum_{t=1}^{T} H_{n,a_n}(f_{\theta_t})
+\lambda_n\mathcal{P}(\boldsymbol{f}_{\boldsymbol{\theta}}) \\
=&-\sum_{t=1}^{T}
\frac{1}{n(n-1)}
\sum_{i\neq j}
I(y_{ti}>y_{tj})\,
s\big(f_{\theta_t}(\boldsymbol{x}_i)-f_{\theta_t}(\boldsymbol{x}_j);a_n\big)
+\lambda_n\sum_{j=1}^{p}\|\mathbf{W}_0(j)\|_2,
\end{split}
\end{equation}
where $\lambda_n$ is a tuning parameter, and $\mathbf{W}_0(j)$ denotes the $j$-th column of the shared first-layer weight matrix $\mathbf{W}_0$, representing the contribution of the $j$-th predictor across all tasks. The group-Lasso penalty $\mathcal{P}(\boldsymbol{f}_{\boldsymbol{\theta}})$ encourages groupwise shrinkage of the first-layer columns and promotes the selection of important predictors shared across tasks. The smoothed rank-based objective provides a unified and scale-comparable criterion across tasks, avoiding the need to manually balance task-specific losses with incompatible magnitudes. Moreover, the criterion remains well defined when the task-specific transformations $\mathcal{D}_t$ are unknown.

\subsection{Implementation}
\begin{figure}
\centerline{
\includegraphics[width=0.8\linewidth]{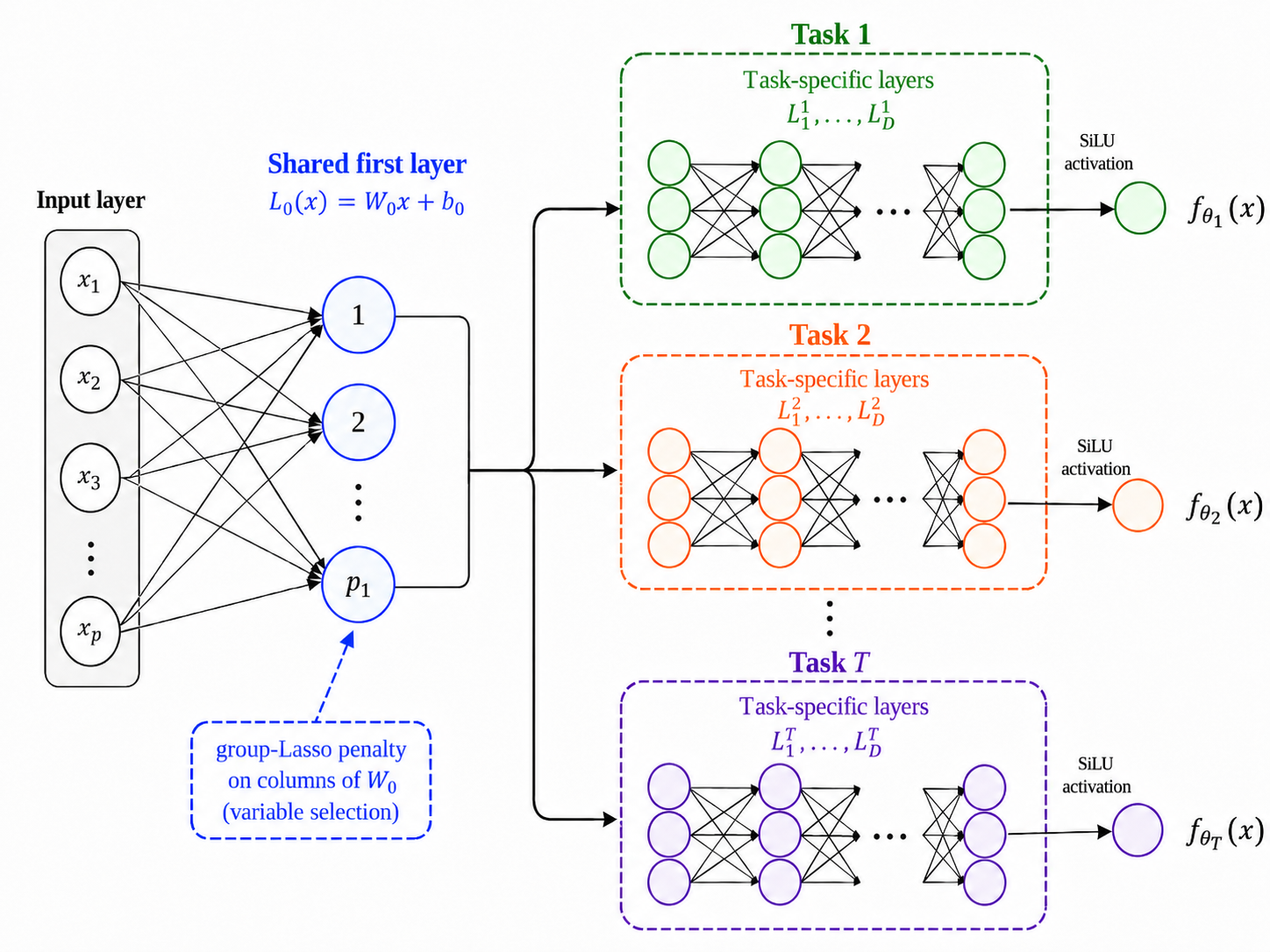}}
\caption{Architecture of multitask deep neural network}
\label{fig:architecture}
\end{figure}

In implementation, we use a multitask neural network $\boldsymbol{f}_{\boldsymbol{\theta}}:\mathbb{R}^p\to\mathbb{R}^T$. For task $t$, the network is defined as
\begin{equation}
f_{\theta_t}(\boldsymbol{x})=\mathcal{L}_D^{\,t}\circ \sigma \circ \mathcal{L}_{D-1}^{\,t}\circ \sigma \circ \cdots \circ \sigma \circ \mathcal{L}_1^{\,t}\circ \sigma \circ \mathcal{L}_0(\boldsymbol{x}),\qquad t=1,\ldots,T,
\label{eq:network}
\end{equation}
where $\sigma(\zeta)=\zeta/(1+e^{-\zeta})$ is the Sigmoid Linear Unit (SiLU) activation function, $\mathcal{L}_0(\boldsymbol{x})=\mathbf{W}_0\boldsymbol{x}+\boldsymbol{b}_0$ is the shared first layer, and $\mathcal{L}_i^t(\boldsymbol{x})=\mathbf{W}_i^t\boldsymbol{x}+\boldsymbol{b}_i^t$ for $i=1,\ldots,D$. Here, $\mathbf{W}_i^t\in\mathbb{R}^{p_{i+1}\times p_i}$ and $\boldsymbol{b}_i^t\in\mathbb{R}^{p_{i+1}}$ are the weight matrix and bias vector of layer $i$, and $p_i$ is the width of the $i$-th layer for a single task. The input dimension is $p_0=p$, and the output dimension is $p_{D+1}=1$ for each task, with the architecture shown in Figure~\ref{fig:architecture}. In addition, we use the SiLU activation function rather than ReLU because SiLU provides a smooth alternative to ReLU, which is especially convenient for the smoothed rank-based objective. 

To reduce the computational complexity inherent in rank-based loss, we train the network using mini-batches and reuse model parameters from the previous epoch. Specifically, at the $(l+1)$-th epoch, the smoothed rank-based loss for the $k$-th batch of task $t$ is approximated as
\begin{equation*}
H_{n,a_n}^{(l+1)}(f_{\theta_t},f_{\theta_t^{(l)}})=\frac{1}{m(n-1)}\sum\limits_{i=1}^{m}\sum_{j=1,j\neq i}^{n}I(\mathit{y}_{ti}^k>\mathit{y}_{tj})s\left(f_{\theta_t}(\boldsymbol{x}_{i}^k)-f_{\theta_t^{(l)}}(\boldsymbol{x}_j);a_n\right),
\end{equation*} 
where $m$ denotes the batch size, $(\boldsymbol{x}_{i}^k,\mathit{y}_{ti}^k)$ denotes the samples in the $k$-th batch, and $f_{\theta_t^{(l)}}$ refers to the model from the $l$-th epoch. The algorithm is summarized in Algorithm~\ref{alg:THNN}.

\begin{algorithm} 
\caption{Deep Smoothed Rank Estimation}\label{alg:THNN}
\begin{algorithmic}[1]
\Require  Data $\{\boldsymbol{x}_i,y_{1i},\cdots,y_{Ti}\}_{i=1}^n$, tuning parameters $\lambda_n$, $a_n$, and total epochs $N_{\rm ep}$
\State Initialize the network $\boldsymbol{f}_{\boldsymbol{\theta}^{(0)}}$ with weights drawn from the uniform distribution on $(-1,1)$
\For {$l=1,\ldots,N_{\rm ep}$}
\State Compute $f_{\boldsymbol{\theta}^{l-1}}(\boldsymbol{x_i})$ for $i=1,\ldots,n$
\For {$k=1,\ldots,\lceil n/m \rceil$\textsuperscript{*}}
\State Update $\theta$ with the AdamW optimizer to minimize the loss
\State $-\sum_{t=1}^{T}H_{n,a_n}^{(l)}(f_{\theta_t},f_{\theta_t^{(l-1)}}  )+\lambda_n\sum_{j=1}^{p}||\mathbf{W}_0^{(l)}(j)||_2$
\EndFor 
\EndFor
\Ensure $\boldsymbol{\hat{f}}_{\boldsymbol{\theta}}=(\hat{f}_{\theta_1},\cdots,\hat{f}_{\theta_T})$
\end{algorithmic}
{\vspace{-1em}\footnotesize{\textsuperscript{*}\ $\lceil \cdot \rceil$ is the ceiling function.}}
\end{algorithm}

\section{Theoretical Properties}\label{sec:theory}
\subsection{Identifiability and Excess Risk Bounds}
We first establish the identifiability of the proposed rank-based criterion. The population versions of the unsmoothed and smoothed rank-based criteria are defined as
\begin{align*}
H_T(\boldsymbol{f})=\sum_{t=1}^{T}H(f_t)=&\sum_{t=1}^{T}\mathbb{E}_{\boldsymbol{X},\widetilde{\boldsymbol{X}},Y_t,\widetilde{Y}_t}[I(\mathit{Y}_{t}>\widetilde{Y}_{t})I(f_t(\boldsymbol{X})>f_t(\widetilde{\boldsymbol{X}}))], \\
H_{a_n,T}(\boldsymbol{f})=\sum_{t=1}^{T}H_{a_n}(f_t)=&\sum_{t=1}^{T}\mathbb{E}_{\boldsymbol{X},\widetilde{\boldsymbol{X}},Y_t,\widetilde{Y}_t}[I(\mathit{Y}_{t}>\widetilde{Y}_{t})s(f_t(\boldsymbol{X})-f_t(\widetilde{\boldsymbol{X}});a_n)],
\end{align*}
where $(\widetilde{\boldsymbol{X}},\widetilde{Y}_t)$ is an i.i.d. copy of $(\boldsymbol{X},Y_t)$. Whenever there is no ambiguity, we omit the subscripts on the expectation operator. We state the main results here and present the technical conditions for identifiability and proofs in the Supplementary Materials.
\begin{proposition}\label{pro:identifiablity}
Under the assumptions in Web Appendix A, the following statements hold.\\
$(i)$ The target function $\boldsymbol{f}^*=(f_1^*,\cdots,f_T^*)$ is the unique maximizer of $H_T(\boldsymbol{f})$ over the function class $\mathcal{F}$ specified in Web Appendix A.\\ 
$(ii)$ For each $a_n>0$, there exists a maximizer $(f^*_{1,a_n},\cdots,f^*_{T,a_n})$ of $H_{a_n,T}(\boldsymbol{f})$ over $\mathcal{F}$, and $\sup\limits_{\boldsymbol{x}}|f^*_{t,a_n}(\boldsymbol{x})-f^*_t(\boldsymbol{x})|\to 0~(t=1,\ldots,T)$, as $a_n\to 0$.
\end{proposition}
Proposition~\ref{pro:identifiablity} provides the population-level justification for the proposed rank-based criterion. Part~$(i)$ establishes $\boldsymbol{f}^*$ as the unique maximizer of the unsmoothed population objective, and Part~$(ii)$ establishes that the maximizer of the smoothed population criterion converges to the same target as the smoothing parameter tends to zero. Thus, the smoothing approximation introduced for computation does not change the population target of estimation. We next formalize the shared sparsity assumption on the target functions. 
\begin{assumption}\label{assum:4}
The support of $X$ is a bounded compact set in $\mathbb{R}^p$. Without loss of generality, we take $\mathcal{X}=[0,1]^p$.
\end{assumption}
\begin{assumption}\label{assum:5}
For each task $t$, there exists a function $\tilde{f}_t^*:\mathcal{X}_s\to\mathbb{R}$ belonging to the Hölder class $\mathcal{H}^{\beta}(\mathcal{X}_s,B_0)$ such that $f_t^*(\boldsymbol{X})=\tilde{f}_t^*(\boldsymbol{X}^s),$
where $\mathcal{X}_s$ denotes the support of $\boldsymbol{X}^s$.
\end{assumption}
Here $\mathcal H^\beta(\mathcal X_s,B_0)$ denotes the usual Hölder ball with smoothness $\beta$ and radius $B_0$, and the formal definition is given in Web Appendix A. Assumptions~\ref{assum:4} and \ref{assum:5} describe a sparse nonparametric setting in which the target functions depend only on a fixed subset of predictors and satisfy a Hölder smoothness condition. Such assumptions are standard in high-dimensional nonparametric analysis \citep{schmidt2020nonparametric,Jiao2023} and provide the basis for the approximation and variable selection arguments developed below.

For simplicity, we assume that the task-specific subnetworks after the shared first layer have a common architecture but task-specific parameters. The analysis can be extended to settings in which the task-specific architectures differ across tasks. Let $\mathcal F_{\boldsymbol{\theta}}=\mathcal F_{D,\widetilde{W},S,B}$ denote the multitask neural-network class defined in~\eqref{eq:network}, where $D$ is the depth, $\widetilde W$ is the maximum hidden-layer width for a single task, $S$ is the number of network parameters for a single task, and $B$ is a uniform bound such that $\|f_{\theta_t}\|_\infty\leq B$ for all $t$. Specifically, for $M,N\in\mathbb{N}^+$, we take
$D=21(\lfloor \beta \rfloor+1)^2M\lceil \log_2(8M)\rceil+2p_s+1$ and $\widetilde{W}=38(\lfloor \beta \rfloor+1)^23^{p_s}p_s^{\lfloor\beta\rfloor+1}N\lceil \log_2(8N)\rceil,$
where $\mathbb{N}^+$ denotes the set of positive integers. The deep smoothed rank estimator is defined as $$\hat{\boldsymbol{\theta}}_{n,T}\in\arg\min_{\boldsymbol{f}_{\boldsymbol{\theta}}\in\mathcal{F}_{\boldsymbol{\theta}}} L_n(\boldsymbol{f}_{\boldsymbol{\theta}}).$$ 
We write $A\lesssim B$, equivalently $A=O(B)$, if $A\leq cB$ for some positive constant $c$, and $A\asymp B$ if $A\lesssim B$ and $B\lesssim A$. We next establish a nonasymptotic error bound for the unsmoothed excess risk  $H_T(\boldsymbol{f}^*)-H_T(\boldsymbol{f}_{\hat{\boldsymbol{\theta}}_{n,T}})$, which quantifies the discrepancy between the estimated functions and the target functions.

\begin{theorem}\label{thm:excess_risk}
Suppose that Assumptions in Web Appendix A and \ref{assum:4}-\ref{assum:5} hold and that $B\ge\max\{B_0,1\}$. If the Frobenius norms of the first-layer weight matrix are uniformly bounded, then the smoothed rank-based estimator $\hat{\boldsymbol{\theta}}_{n,T}$ satisfies
$$\mathbb{E}\{H_T(\boldsymbol{f}^*)-H_T(\boldsymbol{f}_{\hat{\boldsymbol{\theta}}_{n,T}})\}\lesssim T\sqrt{\frac{1}{n}SD\log S(\log n-\log a_n)}+Ta_n^{-1}(MN)^{-2\beta/p_s}+Ta_n\log a_n^{-1}+\lambda_n\sqrt{p}.$$
\end{theorem}

The assumption of uniformly bounded weight norms is commonly used in the analysis of deep neural networks \citep{schmidt2020nonparametric,Jiao2023}. The same bounds can be established for the smoothed excess risk $\mathbb{E}\{H_{a_n,T}(\boldsymbol{f}^*)-H_{a_n,T}(\boldsymbol{f}_{\hat{\boldsymbol{\theta}}_{n,T}})\}$ as shown in Web Appendix C. Theorem~\ref{thm:excess_risk} decomposes the excess risk into four terms. The first term is the stochastic error induced by finite-sample estimation and depends on the network complexity through $S$ and $D$. The second term is the approximation error and reflects the ability of the network class to approximate the sparse Hölder target functions. The third term is the smoothing error introduced by replacing the discontinuous indicator function with the smooth sigmoid approximation. The final term is the regularization error due to the group penalty, which vanishes when $\lambda_n=0$, yielding the corresponding unpenalized bound.

We next balance the network complexity, smoothing level, and penalty parameter to obtain explicit rates for the unsmoothed excess risk. Motivated by the approximation advantages of deep networks over shallow networks \citep{Jiao2023}, we fix the network width and choose the depth to optimize the convergence rate when $p$ diverges.

\begin{corollary}\label{cor:excess_risk}
Under the conditions of Theorem~\ref{thm:excess_risk}, for fixed $N\in\mathbb{N}^+$, the following statements hold.\\
(i) If $p=O(\log^{c_1} n)$ for some $0\leq c_1<\infty$, set $M=O(n^{\frac{p_s}{2\beta+2p_s}})$, $a_n^{-1}=O(n^{\frac{\beta}{2\beta+2p_s}})$, $T=O(1)$, and  $\lambda_n=O(n^{-\frac{\beta}{2\beta+2p_s}-c_3})$ for some $c_3>0$. If $p=O(n^{c_2})$ for some $0<c_2<\frac{p_s}{2p_s+2\beta}$, use the same choices of $M,a_n$ and $T$, and set $\lambda_n=O(n^{-\frac{\beta}{2\beta+2p_s}-\frac{c_2}{2}})$. Then, in both cases,
$$\mathbb{E}\{H_T(\boldsymbol{f}^*)-H_T(\boldsymbol{f}_{\hat{\boldsymbol{\theta}}_{n,T}})\}\lesssim n^{-\frac{\beta}{2\beta+2p_s}}\log^4n.$$
(ii) If $p=O(n^{c_2})$ for some $\frac{p_s}{2p_s+2\beta}<c_2<1$, set $M= O(n^{\frac{(1-c_2)p_s}{2\beta+p_s}})$,
$a_n^{-1}=O(n^{\frac{\beta(1-c_2)}{2\beta+p_s}})$, $T=O(1)$, and $\lambda_n=O(n^{-\frac{\beta(1-c_2)}{2\beta+p_s}-c_2/2})$. Then,
$$\mathbb{E}\{H_T(\boldsymbol{f}^*)-H_T(\boldsymbol{f}_{\hat{\boldsymbol{\theta}}_{n,T}})\}\lesssim n^{-\frac{\beta(1-c_2)}{2\beta+p_s}}\log^4n.$$
\end{corollary}

Corollary~\ref{cor:excess_risk} gives explicit convergence rates for the unsmoothed population rank criterion under suitable choices of network complexity, smoothing level, and penalty parameter. Together with Proposition~\ref{pro:identifiablity}, these bounds justify the proposed estimator as a consistent procedure for estimating the target functions under the multitask transformation model.

\subsection{Variable Selection}
Because neural-network parameters are generally not identifiable, the same function may admit multiple parameter representations. To accommodate this nonidentifiability, we formulate variable-selection consistency relative to an equivalence class of population-optimal parameters. Let $\boldsymbol{f}_{\boldsymbol{\theta}^*}\in\arg\max_{\boldsymbol{f}\in\mathcal{F}_{\boldsymbol{\theta}}}H_{a_n,T}(\boldsymbol{f})$ denote a population-optimal network. Define the corresponding equivalence set $\Theta$ as
$$\Theta=\{\boldsymbol{\theta}:H_{a_n,T}(\boldsymbol{f}_{\boldsymbol{\theta}})=H_{a_n,T}(\boldsymbol{f}_{\boldsymbol{\theta}^*}),\boldsymbol{f}_{\boldsymbol{\theta}}\in\mathcal{F}_{\boldsymbol{\theta}}\}.$$ 
To separate the effects of important and unimportant predictors, we decompose the network parameter vector $\boldsymbol{\theta}$ as $(\boldsymbol{u}(\boldsymbol{\theta}),\boldsymbol{v}(\boldsymbol{\theta}),\boldsymbol{\nu}(\boldsymbol{\theta}))$. Here, $\boldsymbol{u}(\boldsymbol{\theta})\in\mathbb R^{p_1\times p_s}$ contains the first-layer weights associated with the important predictors, $\boldsymbol{v}(\boldsymbol{\theta})\in\mathbb R^{p_1\times p_c}$ contains the first-layer weights associated with the unimportant predictors, and  $\boldsymbol{\nu}(\boldsymbol{\theta})$ collects all remaining parameters. These components are unknown and are estimated through the network estimator. Define the distance from $\boldsymbol{\theta}$ to $\Theta$ by $d(\boldsymbol{\theta},\Theta)=\min\limits_{\tilde{\boldsymbol{\theta}}\in\Theta}||\boldsymbol{\theta}-\tilde{\boldsymbol{\theta}}||_2$. We impose the following condition to relate the population criterion gap to the distance from the equivalence set.
\begin{assumption}\label{assum:analytic}
There exist constants $c>0$ and $\mu>2$, independent of $n$, such that $d(\boldsymbol{\theta},\Theta)^\mu\le c|H_{a_n,T}(\boldsymbol{f}_{\boldsymbol{\theta}^*})-H_{a_n,T}(\boldsymbol{f}_{\boldsymbol{\theta}})|.$    
\end{assumption}
Assumption~\ref{assum:analytic} is a local separation condition that imposes a uniform polynomial growth condition on the smoothed population criterion in a neighborhood of the equivalence set $\Theta$. This condition is satisfied for a fixed analytic network class on a compact parameter space and fixed $a_n$, where a Łojasiewicz-type inequality can provide polynomial control of the distance to its zero-level set \citep{ji1992global}. Closely related separation conditions have been used in deep transformation models \citep{Wang2024deep}, and analogous analytic arguments have been used to study variable selection for neural-network representations \citep{dinh2020consistent}. The following result establishes the convergence of the estimator toward the equivalence set $\Theta$.
\begin{theorem}\label{thm:consistency}
Under the conditions of Theorem~\ref{thm:excess_risk} and Assumption~\ref{assum:analytic}, the smoothed rank-based estimator satisfies 
\begin{align*}
\mathbb{E}d(\hat{\boldsymbol{\theta}}_{n,T},\Theta)
\lesssim& T^{\frac{1}{\mu}}\left[\frac{1}{n}SD\log S(\log n-\log a_n)\right]^{\frac{1}{2\mu}}+(\lambda_n\sqrt{p})^{\frac{1}{\mu}},\\
\mathbb{E}\sum_{j=1}^{p_c}||\boldsymbol{v}(\hat{\boldsymbol{\theta}}_{n,T})(j)||_2\lesssim& \lambda_n^{-1}T\sqrt{\frac{1}{n}SD\log S(\log n-\log a_n)}+\lambda_n^{-1}Ta_n\log a_n^{-1}\\
&+\lambda_n^{-1}Ta_n^{-1}(MN)^{-2\beta/p_s}+\mathbb{E}d(\hat{\boldsymbol{\theta}}_{n,T},\Theta).
\end{align*}
\end{theorem}
Theorem~\ref{thm:consistency} bounds both the distance from the estimator to the equivalence set $\Theta$ and the total first-layer weight norms assigned to unimportant predictors. To gain additional insights, we compare this result with separate single-task estimation, in which the same network architecture and rank-based criterion are applied independently to each task, corresponding to $T=1$ in the proposed framework. Let $\widehat{\theta}_{n}^t$ and $\Theta_t$ denote the single-task estimator and the corresponding equivalence set for task $t$. Applying the same argument to each task gives the aggregate bound
$$\mathbb{E}\sum_{t=1}^{T}d(\hat{\theta}_{n}^t,\Theta_t)\lesssim T\left[{n}^{-1}SD\log S(\log n-\log a_n)\right]^{\frac{1}{2\mu}}+T(\lambda_n\sqrt{p})^{\frac{1}{\mu}}.$$
Compared with this separate single-task bound, the multitask method can yield a tighter bound when $T$ grows with $n$. This reflects the benefit of the shared first layer and the group penalty.

Define $\widehat{\mathcal{A}}_n(\tau_n)=\{j:||\mathbf{W}_{0,\hat{\boldsymbol{\theta}}_{n,T}}(j)||_2>\tau_n\}$ where $\mathbf{W}_{0,\hat{\boldsymbol{\theta}}_{n,T}}(j)$ is the $j$-th column of the first layer weight matrix under $\hat{\boldsymbol{\theta}}_{n,T}$ and $\tau_n<c_0/2$, and $\mathcal{A}^*=\{j:X_j\ \text{is an important predictor}\}$. The following corollary establishes variable-selection consistency.

\begin{corollary}\label{cor:consistent}
Under the conditions of Theorem~\ref{thm:consistency}, if the tuning parameters and the function class $\mathcal{F}_\theta=\mathcal{F}_{D,\widetilde{W},S,B}$ satisfy
\begin{align*}
(\lambda_n\tau_n)^{-1}Ta_n\log a_n^{-1}&\to0,\quad (\lambda_n\tau_n)^{-1}Ta_n^{-1}(MN)^{-2\beta/p_s}\to 0, \quad (\tau_n)^{-1}\lambda_n\sqrt{p}\to0,\\
&(\lambda_n\tau_n)^{-1}T\sqrt{\frac{1}{n}SD\log S(\log n-\log a_n)}\to0  ,  
\end{align*}
as $n\to\infty$, then the smoothed rank estimator $\hat{\boldsymbol{\theta}}_{n,T}$ is variable-selection consistent, that is,
$$\lim\limits_{n\to\infty}\mathbb{P}(\widehat{\mathcal{A}}_n(\tau_n)=\mathcal{A}^*)=1.$$
\end{corollary}

Corollary~\ref{cor:consistent} shows that the proposed estimator consistently identifies the shared important predictors under the stated population separation and tuning conditions. These conditions can be satisfied under suitable choices of the network and tuning parameters. For example, in the logarithmic-dimensional regime of Corollary~\ref{cor:excess_risk}, where $p=O(\log^{c_1} n),T=O(1)$, and $N$ is fixed, we can take $M\asymp n^{\frac{p_s}{2\beta+2p_s}}$, $a_n^{-1}\asymp n^{\frac{\beta}{2\beta+2p_s}}$ and $\lambda_n\asymp n^{-\frac{\beta}{2\beta+2p_s}}\log^5 n,\tau_n\asymp \log^{-\frac{1}{2}} n$.

\section{Simulation}\label{sec:simulation}
We conduct simulation studies to evaluate the performance of the proposed method relative to several competing approaches, including the multitask weighting strategies based on gradient normalization (GradNorm) \citep{chen2018gradnorm}, uniform weighting (Uniform), dynamic weight averaging (DWA) \citep{liu2019end}, and uncertainty weighting (Uncertainty) \citep{kendall2018multi}, and the single-task learning method with the rank-based loss (STL). Given the absence of a unified loss for different outcome types, the competing multitask methods use mean Huber loss for regression tasks and cross-entropy loss for classification tasks, together with the same group penalty for variable selection. Because rank-based estimation identifies the target functions up to monotone transformations, we further apply a calibration step to obtain predicted responses or class labels. To ensure a fair comparison and separate the contribution of the multitask transformation model, all methods use the same network architecture, hyperparameter settings, and data splits. Additional implementation details are provided in Web Appendix D.

In each setting, we generate $n=1000$ samples and randomly split them into training, validation, and testing sets in a $5:3:2$ ratio. We consider predictor dimensions $p\in\{100,200,500,800\}$ and fix the number of important predictors at $p_s=30$. Three multitask settings are examined: a two-task regression setting, a two-task classification setting, and a mixed classification-regression setting. In all settings, the predictors $\boldsymbol{X}$ are generated from $N(\boldsymbol{0}_p,\boldsymbol{\Sigma})$, where $\boldsymbol{\Sigma}=\{0.3^{|k-j|}\}_{k,j=1}^{p}$. Regression errors follow the contaminated distribution $\epsilon_t\sim 0.8N(0,1)+0.2\mathrm{Cauchy}(0,1)$ to induce heavy-tailed noise, and $20\%$ of classification labels are randomly flipped to introduce label noise.

\noindent \textbf{Setting\ 1}: $Y_t=f_t^*(\boldsymbol{X})+\epsilon_t=\boldsymbol{X}^\top\boldsymbol{\beta}_t+\sin(\boldsymbol{X}^\top\boldsymbol{\beta}_t)+\tanh(\boldsymbol{X}^\top\boldsymbol{\beta}_t)+ ((\boldsymbol{X}^\top\boldsymbol{\beta}_t)^2+1)^{-1}+\epsilon_t$, where $\boldsymbol{\beta}_t=(\boldsymbol{\beta}_{p_s,t},\boldsymbol{0}_{p_c})$. The vector $\boldsymbol{\beta}_{p_s,1}$ is $p_s$-dimensional, with entries sampled independently from $U(0.2,0.8)$. The vector $\boldsymbol{\beta}_{p_s,2}$ is constructed by randomly replacing 80\% of the entries in $\boldsymbol{\beta}_{p_s,1}$ with independent draws from $U(1.2,1.8)$. This setting defines a single-index two-task regression problem.

\noindent \textbf{Setting\ 2}: $Y_t=f_t^*(\boldsymbol{X})+\epsilon_t=\left(\boldsymbol{X}_{s_1}^2,(\boldsymbol{X}_{s_2}^2+1)^{-1},\sin\boldsymbol{X}_{s_3},\tanh\boldsymbol{X}_{s_4},\exp(\boldsymbol{X}_{s_5}^2+1)^{-1},\boldsymbol{X}^c\right)^\top\boldsymbol{\beta}_t+\epsilon_t$, where $\boldsymbol{X}^s=(\boldsymbol{X}_{s_1},\boldsymbol{X}_{s_2},\boldsymbol{X}_{s_3},\boldsymbol{X}_{s_4},\boldsymbol{X}_{s_5})$, with each component having dimension $p_s/5$, and $\boldsymbol{\beta}_t$ is constructed as in Setting 1. This setting defines a fully nonlinear two-task regression problem.

\noindent \textbf{Setting\ 3}: $Y_t=I\{f_t^*(\boldsymbol{X})>c_t\}$, where $f_t^*(\boldsymbol{X})$ is defined as in Setting 1, and $c_t$ is a fixed constant chosen in a preliminary pilot simulation to avoid severe class imbalance and kept fixed across all replications. This setting defines a single-index two-task classification problem. 

\noindent \textbf{Setting\ 4}: $Y_t=I\{f_t^*(\boldsymbol{X})>c_t\}$, where $f_t^*(\boldsymbol{X})$ is defined as in Setting 2 and $c_t$ is the fixed threshold described above. This setting defines a fully nonlinear two-task classification problem.

\noindent \textbf{Setting\ 5}: $Y_1=f_1^*(\boldsymbol{X})+\epsilon_1,Y_2=I\{f_2^*(\boldsymbol{X})>c_2\}$
where $f_1^*(\boldsymbol{X})$ is the first task function from Setting 2 and $f_2^*(\boldsymbol{X})$ is the second task function from Setting 4. This setting defines a fully nonlinear mixed classification-regression problem.

\noindent \textbf{Setting\ 6}: $Y_1=f_1^*(\boldsymbol{X})+\epsilon_1,Y_2=I\{f_2^*(\boldsymbol{X})>c_2\}$
where $f_1^*(\boldsymbol{X})$ is the second task function from Setting 2 and $f_2^*(\boldsymbol{X})$ is the first task function from Setting 4. This setting defines a fully nonlinear mixed classification-regression problem.

For classification tasks, predictive performance is evaluated using accuracy, precision, recall, and F1 score. For regression tasks, we report mean absolute error (MAE), root mean squared error (RMSE), mean Cauchy loss (Cauchy), and mean Huber loss (Huber), all computed on the testing set. The definitions of all metrics are given in Web Appendix A. Because the theoretical threshold $\tau_n$ is not directly specified in practice, we assess variable-selection performance using a ranking-based measure. Specifically, we rank predictors by the column norms of the estimated first-layer weight matrix and report the number of true positives (TP), defined as the number of truly important predictors among the top $p_s$ ranked predictors. The performance of the proposed method is evaluated over 100 simulation replications. Results for $p=100$ are reported in Tables \ref{tab:reg_p100} and \ref{tab:mixed_p100}, and additional results in the Supplementary Materials show similar patterns as the feature dimension increases.

For the single-type multitask settings in Table~\ref{tab:reg_p100}, the proposed method generally yields smaller regression errors, especially in terms of MAE, RMSE, and Huber loss, and achieves competitive classification performance with the highest mean accuracy and F1 score. The advantage is more pronounced in the mixed-type setting in Table~\ref{tab:mixed_p100}, where regression and classification tasks are learned jointly. In this setting, the proposed method attains the best classification performance and substantially smaller regression errors than the competing multitask weighting strategies. These findings support the use of a unified scale-comparable rank-based objective, which avoids explicit balancing of task-specific losses defined on incompatible scales. Compared with single-task learning, the multitask estimator generally improves both predictive and variable-selection performance, suggesting that the shared first layer and group penalty effectively exploit the shared sparse predictor structure across tasks. The favorable performance under heavy-tailed regression errors and classification label noise further suggests that the rank-based objective is robust to outcome perturbations in both continuous and binary tasks.

\begin{table}
\centering
\caption{Simulation results for the single-type settings with $p=100$: mean (sd)}
\label{tab:reg_p100}
\resizebox{\textwidth}{!}{
\begin{tabular}{clccccc}
\hline
Setting & Method & MAE & RMSE & Cauchy & Huber & TP \\
\hline
\multirow{6}{*}{1}
& Proposed    & \textbf{2.046(0.432)} & \textbf{5.935(3.549)} & 1.111(0.063) & \textbf{2.016(0.573)} & 29.138(0.868) \\
& GradNorm    & 2.569(2.932) & 16.185(40.285) & \textbf{1.018(0.050)} & 2.744(3.944) & 29.880(0.327) \\
& Uniform     & 2.947(2.932) & 16.410(40.262) & 1.293(0.063) & 3.209(3.944) & 27.250(1.104) \\
& DWA         & 2.572(2.931) & 16.185(40.282) & 1.020(0.050) & 2.747(3.942) & 29.900(0.302) \\
& Uncertainty & 2.571(2.931) & 16.185(40.283) & 1.019(0.049) & 2.745(3.942) & \textbf{29.930(0.256)} \\
& STL         & 3.124(3.162) & 16.639(40.214) & 1.302(0.369) & 3.453(4.251) & 24.955(1.946) \\
\hline
\multirow{6}{*}{2}
& Proposed    & \textbf{2.779(0.303)} & \textbf{5.210(1.776)} & 1.635(0.080) & \textbf{2.957(0.401)} & 18.350(1.600) \\
& GradNorm    & 3.180(1.659) & 10.523(22.678) & 1.659(0.087) & 3.498(2.230) & \textbf{18.590(2.040)} \\
& Uniform     & 4.855(1.689) & 11.988(22.484) & 2.394(0.109) & 5.702(2.271) & 10.710(1.671) \\
& DWA         & 3.107(1.666) & 10.443(22.690) & \textbf{1.630(0.090)} & 3.403(2.241) & 16.420(2.123) \\
& Uncertainty & 3.251(1.653) & 10.583(22.658) & 1.699(0.099) & 3.590(2.221) & 17.880(2.872) \\
& STL         & 7.739(24.782) & 15.508(35.148) & 2.022(0.795) & 9.609(33.322) & 16.880(1.589) \\
\hline
Setting & Method & Accuracy & Precision & Recall & F1 & TP \\
\hline
\multirow{6}{*}{3}
& Proposed    & \textbf{0.711(0.021)} & \textbf{0.762(0.181)} & \textbf{0.999(0.002)} & \textbf{0.727(0.021)} & \textbf{23.575(1.310)} \\
& GradNorm    & 0.705(0.026) & 0.706(0.040) & 0.709(0.047) & 0.705(0.031) & 23.130(2.008) \\
& Uniform     & 0.664(0.026) & 0.665(0.041) & 0.669(0.050) & 0.664(0.032) & 12.750(2.115) \\
& DWA         & 0.706(0.028) & 0.708(0.041) & 0.710(0.049) & 0.707(0.033) & 23.110(2.064) \\
& Uncertainty & 0.707(0.027) & 0.708(0.041) & 0.711(0.045) & 0.708(0.031) & 23.400(1.670) \\
& STL         & 0.669(0.026) & 0.682(0.249) & 0.998(0.003) & 0.698(0.028) & 14.305(3.396) \\
\hline
\multirow{6}{*}{4}
& Proposed    & \textbf{0.599(0.026)} & \textbf{0.716(0.175)} & \textbf{1.000(0.000)} & \textbf{0.662(0.021)} & \textbf{14.450(1.449)} \\
& GradNorm    & 0.595(0.030) & 0.594(0.049) & 0.579(0.064) & 0.582(0.041) & 14.350(1.997) \\
& Uniform     & 0.556(0.027) & 0.551(0.045) & 0.555(0.066) & 0.549(0.038) & 9.820(2.115) \\
& DWA         & 0.593(0.029) & 0.592(0.047) & 0.578(0.072) & 0.580(0.043) & 14.210(2.051) \\
& Uncertainty & 0.592(0.032) & 0.593(0.053) & 0.573(0.085) & 0.576(0.050) & 14.390(1.853) \\
& STL         & 0.569(0.026) & 0.554(0.252) & 0.999(0.003) & 0.646(0.029) & 10.930(1.870) \\
\hline
\end{tabular}
}
\end{table}

\begin{table}
\centering
\caption{Simulation results for the mixed-type setting with $p=100$: mean (sd)}
\label{tab:mixed_p100}
\resizebox{\textwidth}{!}{
\begin{tabular}{clccccccccc}
\hline
Setting & Method & Accuracy & Precision & Recall & F1 & MAE & RMSE & Cauchy & Huber & TP \\
\hline
\multirow{6}{*}{5}
& Proposed    & \textbf{0.627(0.004)} & \textbf{0.830(0.067)} & \textbf{1.000(0.000)} & \textbf{0.680(0.003)} & \textbf{2.051(0.284)} & \textbf{3.892(0.396)} & 1.276(0.152) & \textbf{2.006(0.368)} & 17.512(1.534) \\
& Uniform     & 0.556(0.018) & 0.557(0.017) & 0.561(0.039) & 0.557(0.020) & 3.438(2.442) & 11.803(2.392) & 1.819(1.028) & 3.819(2.617) & 11.430(2.212) \\
& DWA         & 0.592(0.036) & 0.593(0.038) & 0.625(0.071) & 0.577(0.037) & 2.472(2.433) & 11.095(2.376) & \textbf{1.220(0.928)} & 2.582(2.608) & \textbf{17.730(2.278)} \\
& Uncertainty & 0.588(0.033) & 0.567(0.070) & 0.597(0.095) & 0.552(0.075) & 2.530(2.428) & 11.131(2.356) & 1.260(0.928) & 2.658(2.613) & 15.650(2.886) \\
& GradNorm    & 0.615(0.031) & 0.622(0.029) & 0.595(0.085) & 0.605(0.054) & 2.545(2.436) & 11.151(2.358) & 1.268(0.927) & 2.676(2.621) & 16.280(1.970) \\
& STL         & 0.563(0.016) & 0.533(0.060) & 0.998(0.004) & 0.654(0.012) & 2.943(0.560) & 11.536(1.217) & 1.449(0.241) & 3.192(0.744) & 13.840(2.522) \\
\hline
\multirow{6}{*}{6}
& Proposed    & \textbf{0.626(0.004)} & \textbf{0.832(0.067)} & \textbf{1.000(0.000)} & \textbf{0.680(0.003)} & \textbf{3.738(0.457)} & \textbf{5.566(0.615)} & \textbf{2.183(0.245)} & \textbf{4.199(0.588)} & \textbf{18.262(1.482)} \\
& Uniform     & 0.556(0.015) & 0.553(0.017) & 0.555(0.036) & 0.552(0.018) & 6.666(2.278) & 13.856(2.250) & 3.091(0.979) & 8.105(2.468) & 10.590(1.821) \\
& DWA         & 0.592(0.029) & 0.577(0.064) & 0.564(0.095) & 0.545(0.067) & 4.341(2.244) & 11.561(2.247) & 2.270(0.810) & 5.007(2.435) & 15.580(3.376) \\
& Uncertainty & 0.613(0.033) & 0.614(0.034) & 0.583(0.087) & 0.590(0.056) & 4.353(2.234) & 11.585(2.237) & 2.278(0.815) & 5.024(2.440) & 13.030(2.500) \\
& GradNorm    & 0.604(0.030) & 0.608(0.028) & 0.569(0.101) & 0.579(0.064) & 4.765(2.323) & 12.032(2.347) & 2.419(0.877) & 5.572(2.553) & 15.050(2.587) \\
& STL         & 0.565(0.013) & 0.615(0.066) & 0.998(0.004) & 0.645(0.014) & 4.604(0.399) & 11.855(0.956) & 2.358(0.229) & 5.357(0.559) & 14.475(2.193) \\
\hline
\end{tabular}
}
\end{table}

\section{Data Analysis}\label{sec:real_data}
Biomedical studies increasingly collect multiple related outcomes of different types from the same subjects, including continuous molecular traits, binary disease indicators, and clinically defined subtypes. In cancer genomics, for example, gene-expression profiles are often used to study disease status, inflammatory biomarkers, and molecular subtypes. These outcomes are measured on different scales but may reflect common tumor-related regulatory mechanisms. Such settings motivate the joint analysis of multiple outcomes using deep learning methods that can capture nonlinear gene–outcome associations while exploiting information through shared molecular structures.

We evaluate the proposed method using three datasets: mouse genetics, METABRIC, and lung cancer data, and we compare it with the same competing approaches used in the simulation study. For each dataset, samples are randomly split into training and testing sets in a $7:3$ ratio, and predictive performance is evaluated over $100$ independent random splits. Predictive performance results are summarized in Tables~\ref{tab:real_single} and~\ref{tab:real_mixed}, and variable-selection results are provided in Web Appendix D. 

The mouse genetics dataset, obtained from the Gene Expression Omnibus under accession number GSE3330, contains gene-expression measurements for $22,575$ genes in $60$ mice from two inbred mouse populations. We analyze two continuous physiological phenotypes, stearoyl-CoA desaturase 1 (SCD1) and phosphoenolpyruvate carboxykinase (PEPCK), yielding a two-task regression setting. These phenotypes are involved in lipid synthesis and glucose homeostasis, and their dysregulation is associated with hepatic steatosis, obesity, and insulin resistance. Identifying genes jointly associated with SCD1 and PEPCK may reveal shared regulatory mechanisms linking lipid and glucose metabolism.

As shown in Table~\ref{tab:real_single}, the proposed method achieves the best predictive performance, with the lowest MAE, RMSE, Cauchy loss, and Huber loss. Compared with the closest competing method, GradNorm, the proposed method reduces prediction errors and yields smaller standard deviations across all criteria. These results suggest that the unified rank-based multitask objective improves both prediction and stability. The selected genes are also biologically interpretable. In particular, \textit{Scd1} is directly involved in lipid metabolism, hepatic steatosis, obesity, and insulin sensitivity \citep{miyazaki2007hepatic}, and \textit{Sult2a2} belongs to the murine hepatic SULT2A family, whose members play important roles in hepatic steroid and xenobiotic metabolism \citep{kocarek2008sult2a}.

In the METABRIC study, the Molecular Taxonomy of Breast Cancer International Consortium profiled $1,980$ primary breast cancer samples. Following the established studies \citep{falzone2020identification}, we use $482$ genes and $6$ clinical variables as predictors. Breast cancer is molecularly heterogeneous, and molecular subtype is closely associated with prognosis, treatment response, and targeted therapy selection. We consider four breast cancer subtypes, LumA, LumB, HER2, and Basal, and group them into two binary classification tasks: one separates hormone-receptor-positive subtypes from the remaining subtypes, and the other groups subtypes according to clinical actionability. 

As reported in Table~\ref{tab:real_single}, the proposed method achieves the highest precision ($0.906$), near-perfect recall ($0.999$), and the best F1 score ($0.711$), compared with the corresponding competing values of $0.839$, $0.999$, and $0.707$. Although its accuracy ($0.779$) is slightly lower than the best accuracy among the weighting-based methods ($0.795$), its superior F1 score indicates a more favorable balance between false-positive and false-negative errors. The leading selected genes include \textit{E2F1}, a cell-cycle transcription factor associated with breast cancer metastasis and altered cell migration \citep{hollern2019e2f1}, and \textit{HER4}, which has been reported as a prognostic marker in breast cancer \citep{wang2016her4}.

The lung cancer dataset, obtained from the Gene Expression Omnibus under accession number GSE252168, contains blood-based gene-expression measurements for $30,715$ genes from $234$ samples, including patients with lung cancer at St. Olavs University Hospital and control subjects from two biobanks. Lung cancer is biologically heterogeneous, and blood-based molecular measurements may provide a minimally invasive source of diagnostic and monitoring information. Beyond distinguishing cancer cases from controls, jointly modeling cancer status with C-reactive protein (CRP), a marker of systemic inflammation, may help identify genes linking disease status to the accompanying inflammatory response. We consider two responses: a binary case-control indicator and the continuous CRP level, forming a mixed classification-regression setting.

As shown in Table~\ref{tab:real_mixed}, the proposed method delivers the strongest overall performance. It achieves the highest accuracy, precision, recall, and F1 score ($0.968$, $0.990$, $0.997$, $0.968$), exceeding single-task learning, the second-best method ($0.951$, $0.969$, $0.989$, $0.953$). For CRP prediction, the proposed method also yields the lowest MAE, RMSE, and Huber loss ($18.803$, $35.315$, $24.365$), improving substantially over the second-best method ($19.308$, $40.694$, $25.113$). Compared with competing multitask weighting methods, it provides a more stable balance between classification and regression performance, suggesting that the unified rank-based objective can handle mixed-type outcomes without relying on potentially unstable combinations of task-specific losses. Among the leading selected genes, \textit{RPLP2} has been reported to be overexpressed in lung adenocarcinoma and associated with clinical outcomes \citep{xu2024rplp2}, whereas \textit{HBB} has been proposed as a biomarker for lung cancer diagnosis and monitoring \citep{xu2024hbb}.

Taken together, the three data analyses show that the proposed method performs favorably across a variety of multitask settings. Compared with the multitask weighting methods, these results demonstrate the advantage of replacing weighted task-specific losses with a unified and scale-comparable rank-based objective. The improvements over the single-task learning further highlight the value of borrowing information across related biological outcomes. In addition, the gene-identification results show that the proposed method can detect biologically meaningful predictors shared across related outcomes, helping link predictive performance to common molecular mechanisms.

\begin{table}
\centering
\caption{Data analysis results for the single-type multitask settings: mean (sd)}
\label{tab:real_single}
\resizebox{0.9\textwidth}{!}{
\begin{tabular}{llcccc}
\hline
Dataset & Method & MAE & RMSE & Cauchy & Huber \\
\hline
\multirow{6}{*}{Mouse genetics}
& Proposed    & \textbf{0.827(0.086)} & \textbf{1.054(0.112)} & \textbf{0.551(0.072)} & \textbf{0.506(0.085)} \\
& GradNorm    & 0.865(0.130) & 1.163(0.188) & 0.571(0.104) & 0.576(0.146) \\
& Uniform     & 1.772(0.958) & 2.016(0.912) & 1.234(0.628) & 1.652(1.207) \\
& DWA         & 0.923(0.162) & 1.223(0.196) & 0.619(0.134) & 0.632(0.170) \\
& Uncertainty & 0.909(0.172) & 1.189(0.211) & 0.609(0.143) & 0.612(0.188) \\
& STL         & 0.924(0.150) & 1.169(0.187) & 0.630(0.124) & 0.612(0.168) \\
\hline
Dataset & Method & Accuracy & Precision & Recall & F1 \\
\hline
\multirow{6}{*}{METABRIC}
& Proposed    & 0.779(0.012) & \textbf{0.906(0.056)} & \textbf{0.999(0.001)} & \textbf{0.711(0.013)} \\
& GradNorm    & 0.794(0.012) & 0.792(0.022) & 0.640(0.026) & 0.706(0.018) \\
& Uniform     & 0.780(0.014) & 0.750(0.025) & 0.654(0.027) & 0.697(0.020) \\
& DWA         & \textbf{0.795(0.013)} & 0.795(0.022) & 0.640(0.029) & 0.707(0.019) \\
& Uncertainty & \textbf{0.795(0.013)} & 0.795(0.019) & 0.641(0.028) & 0.707(0.019) \\
& STL         & 0.772(0.016) & 0.839(0.151) & \textbf{0.999(0.002)} & 0.699(0.020) \\
\hline
\end{tabular}
}
\end{table}

\begin{table}
\centering
\caption{Data-analysis results for the mixed-type multitask setting: mean (sd)}
\label{tab:real_mixed}
\resizebox{\textwidth}{!}{
\begin{tabular}{llcccccccc}
\hline
Dataset & Method & Accuracy & Precision & Recall & F1 & MAE & RMSE & Cauchy & Huber \\
\hline
\multirow{6}{*}{Lung cancer}
& Proposed    & \textbf{0.968(0.024)} & \textbf{0.990(0.019)} & \textbf{0.997(0.010)} & \textbf{0.968(0.024)} & \textbf{18.803(2.503)} & \textbf{35.315(5.235)} & 4.296(0.472) & \textbf{24.365(3.326)} \\
& GradNorm    & 0.889(0.134) & 0.887(0.158) & 0.949(0.047) & 0.906(0.086) & 19.308(3.513) & 40.694(7.621) & \textbf{3.685(0.363)} & 25.113(4.726) \\
& Uniform     & 0.744(0.216) & 0.630(0.456) & 0.604(0.432) & 0.605(0.428) & 19.645(3.848) & 41.287(7.829) & 3.702(0.350) & 25.562(5.170) \\
& DWA         & 0.748(0.210) & 0.811(0.336) & 0.644(0.430) & 0.629(0.388) & 19.462(3.632) & 40.959(7.775) & 3.720(0.311) & 25.305(4.879) \\
& Uncertainty & 0.868(0.145) & 0.939(0.132) & 0.824(0.274) & 0.833(0.246) & 19.437(3.463) & 40.769(7.936) & 3.764(0.350) & 25.276(4.652) \\
& STL         & 0.951(0.052) & 0.969(0.051) & 0.989(0.019) & 0.953(0.047) & 22.960(9.733) & 60.861(55.131) & 4.472(0.586) & 29.973(13.051) \\
\hline
\end{tabular}
}
\end{table}

\section{Discussion}\label{sec:discussion}
We have developed a multitask transformation framework and a rank-based estimation procedure for mixed-type outcomes under shared sparsity. The proposed approach defines a common estimation objective for mixed-type outcomes rather than aggregating task-specific loss functions. The numerical studies show that the method can provide substantial advantages in some settings, offering a useful alternative to weighted loss aggregation. The theoretical results extend the statistical understanding of multitask learning to nonlinear deep learning models and provide insights for future methodological development. In addition, our work extends the maximum rank correlation idea from single-outcome regression to multitask learning with mixed-type outcomes.

Future work could extend the proposed method in several directions. First, because the rank-based criterion uses only ordering information, it may be less efficient than fully likelihood-based methods, although no appreciable efficiency loss was observed in our numerical studies. Second, the framework could be extended to ordinal, count, censored, and longitudinal responses, for which ties, censoring, or within-subject dependence require additional methodological development. Third, the shared-sparsity assumption could be relaxed to allow both shared and task-specific sparsity patterns, thereby improving robustness to model misspecification and broadening the applicability of the proposed framework.

\section*{Funding}
This work was partially supported by the National Natural Science Foundation of China No. 12571298, Fundamental Research Funds for the Central Universities, Fundamental, Interdisciplinary Disciplines Breakthrough Plan of the Ministry of Education of China (JYB2025XDXM612), and NIH CA204120.\vspace*{-8pt}

\bibliographystyle{biom} \bibliography{reference}

\end{document}